\documentclass{article} 
\usepackage{iclr2020_conference,times}

\usepackage{amsmath,amsfonts,bm}

\def\eqref#1{equation~\ref{#1}}

\def\1{\bm{1}}

\def\rvd{{\mathbf{d}}}
\def\rve{{\mathbf{e}}}

\def\rvk{{\mathbf{k}}}

\def\rvp{{\mathbf{p}}}
\def\rvq{{\mathbf{q}}}

\def\rvs{{\mathbf{s}}}

\def\rvv{{\mathbf{v}}}
\def\rvw{{\mathbf{w}}}

\def\rvy{{\mathbf{y}}}
\def\rvz{{\mathbf{z}}}

\DeclareMathAlphabet{\mathsfit}{\encodingdefault}{\sfdefault}{m}{sl}
\SetMathAlphabet{\mathsfit}{bold}{\encodingdefault}{\sfdefault}{bx}{n}

\def\gL{{\mathcal{L}}}

\def\sD{{\mathbb{D}}}

\usepackage{soul}

\usepackage{tabularx}
\usepackage{hyperref}
\usepackage{url}
\usepackage{graphicx}
\usepackage[export]{adjustbox}
\usepackage{todonotes}
\usepackage{subcaption}
\newcommand{\citepara}[1]{(\cite{#1})}
\usepackage{xspace}
\newcommand{\gpttwo}{GPT-2\xspace}
\usepackage{adjustbox}
\usepackage{multirow}
\usepackage{array,ragged2e}
\newcolumntype{U}{>{\arraybackslash}p{12cm}}
\newcolumntype{Q}{>{\arraybackslash}p{24cm}}
\usepackage{comment}

\title{Style Example-Guided Text Generation using Generative Adversarial Transformers}

\iclrfinalcopy

\author{Kuo-Hao Zeng\thanks{The work was done during internship at NVIDIA.} , Mohammad Shoeybi \& Ming-Yu Liu \\
\texttt{khzeng@cs.washington.edu, \{mshoeybi, mingyul\}@nvidia.com} \\
NVIDIA, Santa Clara, California
}

\begin{document}

\maketitle

\begin{abstract}
	We introduce a language generative model framework for generating a styled paragraph based on a context sentence and a style reference example. The framework consists of a style encoder and a texts decoder. The style encoder extracts a style code from the reference example, and the text decoder generates texts based on the style code and the context. We propose a novel objective function to train our framework. We also investigate different network design choices. We conduct extensive experimental validation with comparison to strong baselines to validate the effectiveness of the proposed framework using a newly collected dataset with diverse text styles. Both code and dataset will be released upon publication. 
\end{abstract}

\section{Introduction}

Text generation refers to the task of generating a new text based on some user input. The task exists in many forms, but arguably the most common form concerns generating a coherent and consistent text based on an input context such as the first few sentences of the target output. This is often achieved by giving the context to a generative language model. Generative language models play a central role in machine learning and natural language processing (NLP). Not only they serve as the main mean for unsupervised feature representation learning but also find use in various applications, including question answering, dialogue agents, summarization, and content creation systems.

Thanks to the introduction of novel deep learning architectures and the availability of large-scale training corpora, the state-of-the-art text generation has advanced significantly in recent years. We can now train language models capable of generating fluent and coherent texts that people cannot tell them apart from those written by humans. However, despite the great achievement, existing generative models are limited and inflexible in the sense that a trained model is only capable in generating texts of one style. It can not be used to generate texts of different styles. For instance, a news generative model can only be used to generate news, and a lyric generative model can only be used to generate lyrics. In contrast, humans can compose texts in various styles.

To bridge the gap, we propose a style example-guided text generation framework that can generate styled texts based on the style of the example reference text. In our framework, the generator takes two inputs where one is the context input while the other is the style reference example. We use the style reference example to change the generation behavior of our generative model dynamically. For a fixed context, when the provided style reference is a paragraph sampled from a news article, it becomes a news generator. When the provided style reference is a review, it becomes a review generator. In other words, the same generator can generate texts of different styles based on the examples. In Tab.~\ref{tab:taser}, we show example outputs of the proposed framework where we generate texts of dramatically different styles for the same input sentence.

\begin{table}[!t]
\begin{adjustbox}{width=1\columnwidth,center}
	\centering
	\begin{tabular}{| l | l |}
	    \cline{1-2}
	    \multicolumn{2}{|l|}{\begin{tabular}{Q}\\
	    {\bf Context -- Wiki Style (Machine Learning Intro.)}\\
Machine learning (ML) is the scientific study of algorithms and statistical models that computer systems use to perform a specific task without using explicit instructions, relying on patterns and inference instead. It is seen as a subset of artificial intelligence. Machine learning algorithms build a mathematical model based on sample data, known as "training data", in order to make predictions or decisions without being explicitly programmed to perform the task.[1][2]:\\
\\
	    \end{tabular}}\\
	    \hline
	    \begin{tabular}{U}
	    {\bf Reference -- News Style}\\
(CNN)Remember that robot dog that went viral a few years ago? The one that can run uphill, climb stairs, and looks like the killer robot dogs in the TV show "Black Mirror"? It's now available for select customers.\\
\\
Boston Dynamics, the company behind the dog, which is named Spot, released a video on Tuesday advertising its launch. It shows Spot navigating challenging terrain, picking up construction objects, moving through bad weather, and picking itself up after a fall.
The video and the product launch page also show off Spot's capabilities -- for instance, it can run at 1.6 meters per second (about 5.2 feet per second), has 360-degree cameras that give it a full range of vision, is dust- and water-proof, and can operate in temperatures ranging from -20 to 45 degrees Celsius (-4 to 113 Fahrenheit).\\ \\ \\
	    \end{tabular}
	    &
	    \begin{tabular}{U}\\
	    {\bf Reference -- Yelp Reviews Style }\\
	    Yumm\\
\\
The line was AWFUL! Came by Saturday around 5:20pm and the line was CRAZY. I do feel like they're understaffed. Despite that, it was my first time and i got the chance to taste test two flavors, matcha and milk tea! Both were great!!\\
\\
I got the :\\
- Matcha swirl with a custard fish, sprinkles and coconut flakes : The custard was so good! Paired great with the ice cream.\\
\\
Staff were so friendly! Especially the cashier (guy) and person who swirled the ice cream that shift! She was super patient as I, myself is very indecisive, haha! Most definitely coming by again!!\\
\\
	    \end{tabular}\\
	    \hline
	    \begin{tabular}{U}\\
	    {\bf Generated}\\
Machine Learning is an ancient concept that has entered the mainstream today, which means that it can survive in the face of such heightened competition. In its application, Flurry wants to improve the efficiency of AI driven computing machines, and with a large footprint. As is seen on me, Flurry has designed a large extension that (unfortunately) can be used in its development, but with lower speed lag. Now that the hybrid technology is here, it's there for the taking.\\
\\
There may also be a Something About Mary in the future. "So it's nice to be able to mount a cat across your desk without requiring any backup access."\\
\\
	    \end{tabular}
	    &
	    \begin{tabular}{U}
	    {\bf Generated}\\
Since the day machines started boosting up, dozens of companies have recently done so. In fact, I highly recommend RoboSports, an example of small-batch ice cream places that have taken up artificial advances in AI; their ice cream is the perfect size, slim and hot, it was great. If you are a connoisseur of ice cream, check out this wonderful ice cream shop.\\ \\ \\ \\ \\
	    \end{tabular}\\
	    \hline
	    
	\end{tabular}
	\end{adjustbox}
	\vspace{-3mm}
	\caption{\small Given the same context sentences but different style reference examples, our generation model can generate paragraphs of different styles. The output paragraph shares the same style as the reference example.}
	\label{tab:taser}
	\vspace{-4mm}
\end{table}

The proposed style example-guided text generation framework is based on the generative adversarial networks (GANs), and we utilize the transformer in both the generator and discriminator design. We collect a large dataset containing documents with many different styles for training. Using a novel learning objective function, our network learns to generate styled texts based on the input style example in an unsupervised manner. We conduct extensive experimental validations with comparisons to strong baselines. We also investigate different ways of designing the generator and compare their performance. Through detailed quantitative and user study results, we prove the effectiveness of the proposed framework for the style example-guided text generation task.

\section{Related Work}

{\bf Language modeling} has seen many advancements in recent years, which has resulted in significant improvements on various NLP tasks. Early language models focused on using n-grams to represent a text distribution. \cite{bengio2003neural} introduced a neural language model in a shift from more traditional n-gram models. Many works later \citepara{mikolov2013distributed, Pennington2013GloVe} focused on word embeddings as a way to represent tokens within the text. More recently, \cite{peters2018deep} used bi-directional LSTMs to obtain deep contextualized word representation. However, RNNs can only represent a limited context. \cite{vaswani2017attention} introduced the transformer networks which use the connections between long-distance word pairs embedded in attention mechanisms and can easily enable the learning of long-term dependency. Many later models \citepara{devlin2018bert, liu2019roberta, dai2019transformer, yang2019xlnet} used transformer model and obtained significant improvements on downstream tasks \citepara{wang2018glue, rajpurkar2016squad, zellers2018swagaf}. Lately, \citepara{radford2019language} introduced \gpttwo, a generative left-to-right language model based on the transformer and showed that these models are able to generate coherent text when pretrained on a large corpus. \cite{megatron} further scaled up the \gpttwo model and demonstrated improved performance. Our work differs from the prior works because we aim for allowing user flexible control over the style of the generated text.

{\bf Texts generation} includes review generation \citepara{radford2017learning, zang2017towards}, sentiment texts generation \citepara{wang2018sentigan, hu2017toward, merity2016pointer}, Wikipedia generation \citepara{liu2018generating, lebret2016neural}, fake news generation \citepara{bakhtin2019real, zellers2019neuralfakenews}, abstractive summarization \citepara{li2018guiding, zhang2019pretraining, pasunuru2017towards}, and conversation/dialogue system \citepara{vinyals2015neural, budzianowski2019hello}. Although many of them trained a transformer on large-scale corpora, their results were limited in their specific domain (e.g., reviews, news, etc.) because they either utilized domain-specific priors in their model design or were not designed to generate texts in many different domains or styles.  

{\bf Control on texts generation}. In addition, there are literature utilizing insertion-base~\citepara{stern2019insertion, chan2019kermit}, GAN-based~\citepara{yu2017seqgan, d2019training}, variational autoencoder-based~\citepara{xu2019unsupervised}, normalizing flow-based~\citepara{tran2019discrete} approaches for general texts generation task. However, we instead focus on generating styled paragraphs conditioning on a context and a reference paragraph. A recent work by~\cite{Keskar2019languagegeneration} is most related to ours. They propose a conditional transformer using a control code to perform language generation in a sequence-to-sequence manner. We demonstrate our method outperforms theirs by a large margin in the experiment section.

{\bf Text style transfer} concerns transferring an input text of one style to a different style~\citepara{ker1992style,rao2018dear,xu2017shakespeare,xu2012paraphrasing,fu2018style,hu2017toward,prabhumoye2018style,shen2017style,dianqiemnlp19}. Our work is different since we do not aim for changing the style of a given text. Instead, we aim for a style-controllable way for generating texts from scratch. Also, rather than handling transferring between two styles (e.g., positive $\leftrightarrow$ negative sentiments), our model can generate texts of many different styles. Finally, our model outputs paragraphs while existing text style transfer works mostly output sentences.  

{\bf Image Style transfer} is a popular topic in computer vision. There are many successful techniques, including iterative optimization on the gram matrix (\cite{gatys2016image}), perceptual loss (\cite{johnson2016perceptual, gupta2017characterizing}), feature transformation (\cite{li2017universal}), adaptive instance-normalization (\cite{dumoulin2016learned, huang2017arbitrary}), and GAN-based methods (\cite{zhu2017unpaired,kim2017learning}). Our proposed framework also gets inspiration from them.
\section{Preliminaries}

Our framework is based on the transformer network~\citepara{vaswani2017attention} and the GAN framework~\citepara{goodfellow2014generative}. In this section, we briefly review these two components.
  
{\bf Transformer} is the state-of-the-art network for various natural language processing tasks. Different from RNNs~\citepara{hochreiter1997long,bengio2003neural,chung2014empirical}, which consume a sequence token by token, in a transformer network, the entire sequence is fed into layers of transformer modules. The representation of a token at a layer is then computed by attending to the latent representations of all the other tokens in the preceding layer.

Variants of transformer networks are available. We build our model based on \gpttwo transformer network~\citepara{radford2019language, megatron}, which train a deep transformer using a left-to-right language model:
\begin{equation}
p(\textbf{w}) = \prod_{t}^{T}p(w_{t}|w_{t-1} ... w_1),
\label{equ:LM}
\end{equation}
where $w_t$'s denote the word tokens. Different from BERT-like transformer networks~\citepara{devlin2018bert, liu2019roberta}, \gpttwo makes a casual assumption, i.e., the latent representation of a token is calculated using only the latent representations of the preceding tokens. Thus, during generation, \gpttwo can be directly applied to complete the text given the context sentence.

{\bf GAN} defines a zero-sum game played by a generator $F$ and a discriminator $D$. Under some nice conditions, the generator learns to convert a random noise vector to a realistic signal in a way that the discriminator cannot tell it apart from real signals. In this case, the distribution of the output signals produced by the generator converges to the distribution of signals observed in the real world.

We use a conditional GAN where $F$ takes a context sentence and a style reference example as inputs. To avoid non-differentiability in text decoding (e.g., beam search), we use a latent GAN formulation~\citepara{achlioptas2017learning}. We first divide $F$ into a feature extractor $F_f$ and an output embedding layer $F_o$; that is $F\equiv F_o \circ F_f$. Now, instead of using the output text from $F_o$ as the discriminator input, we feed the latent representation computed by $F_f$ to the discriminator. For real text, we use a pretrained trained \gpttwo model $H$. Again, we decompose $H$ into a feature extractor $H_f$ and an output embedding layer $H_o$ ($H\equiv H_o \circ H_f$). The GAN discriminator then takes features extracted by $H_f$ as input for real texts. Using this latent GAN formulation, we aim for aligning the feature distribution of our generator to the feature distribution of the pretrained \gpttwo model.

\section{Style Example-Guided Text Generation}

We propose a language generative model framework that allows us to control style of the output text using a style reference example. Given few context sentences $\rvw = \{w_{t}\}_{t=1}^{T}$ and a reference text $\rvs$, our generator $F$ generates output text $\rvy$ that has the same style as the reference example $\rvs$ given by
\begin{equation}
\rvy = F(\rvw, \rvs) \equiv F_o(F_f(\rvw, \rvs)).
\end{equation}
We divide the feature extractor $F_f$  into a style encoder $F_s$ and a text decoder $F_g$ where the style encoder extracts a style representation from the style example, $\rvz=F_s(\rvs)$, and the text decoder $F_g$ consumes the style representation and the context sentences to compute a feature for $F_o$ to generate the styled text  $\rvy$. In this section, we will first introduce the data streams employed during training and our novel learning objective function. We will then discuss various generator design choices.

\subsection{Learning Data Streams}

Let $\sD=\{(\rvd_n, l_n)\}$ be a dataset of documents where $\rvd_n$ is a document and $l_n$ is its style label. We assume a finite set of style labels $\gL=\{1,2,...,L\}$ where each integer represents a style class such as news, review, lyric, poem, novel, and children book. During training, our framework employs two data streams where the first one is called the reconstruction stream while the other is referred to as the cross-style generation stream. We note that such a two-stream processing pipeline is common in GAN-based image translation frameworks~\citepara{liu2017unsupervised,huang2018multimodal,liu2019few} but is less explored for language modeling.

{\bf Reconstruction stream} ($\texttt{RS}$). For this steam, we first sample two documents with the same style from~$\sD$: $(\rvd_i, l_i)$ and $(\rvd_j, l_j)$ where $l_i=l_j$. We then sample two paragraphs\footnote{For the purpose of data augmentation, in our implementation, a paragraph we sample may not be the full paragraph in the nominal sense. It could starting from the middle of a nominal paragraph.}: $\rvp_i\sim\rvd_i$ and $\rvp_j\sim\rvd_j$. We extract the first few sentences from $\rvp_i$ as the input context $\rvw=\psi(\rvp_i)$, where $\psi$ is the extraction function, and use $\rvp_j$ for the style reference $\rvs$. Feeding $\rvw$ and $\rvp_j$ to the generator $F$, we expect $F$ should be able to reconstruct $\rvp_i$: $F(\psi(\rvp_i), \rvp_j) \approx \rvp_i$. 

{\bf Cross-style generation stream} ($\texttt{CS}$). We first sample two documents $(\rvd_i, l_i)\sim\mathcal{D}$ and $(\rvd_k, l_k)\sim\mathcal{D}$ where $l_i\neq l_k$. We then sample paragraphs $\rvp_i\sim\rvd_i$ and $\rvp_k\sim\rvd_k$. We again extract the first few sentences from $\rvp_i$ as the input context $\rvw=\psi(\rvp_i)$ and use $\rvp_k$ for the style reference $\rvs$. As feeding $\rvw$ and $\rvp_k$ to the generator $F$, we expect $F$ should output $\rvp_{i\rightarrow k}=F(\psi(\rvp_i), \rvp_k)$ where $\rvp_{i\rightarrow k}$ should has the same style as $\rvd_k$. Let $C^*$ be an oracle style comparator function that outputs 1 if the two input texts have the same style and 0 otherwise. We aim for  $C^*(\rvp_{i\rightarrow k}, \rvp_k)=1$.

\subsection{Learning Objective}
\begin{figure}[!t]
    \centering
    \includegraphics[width=.99\linewidth]{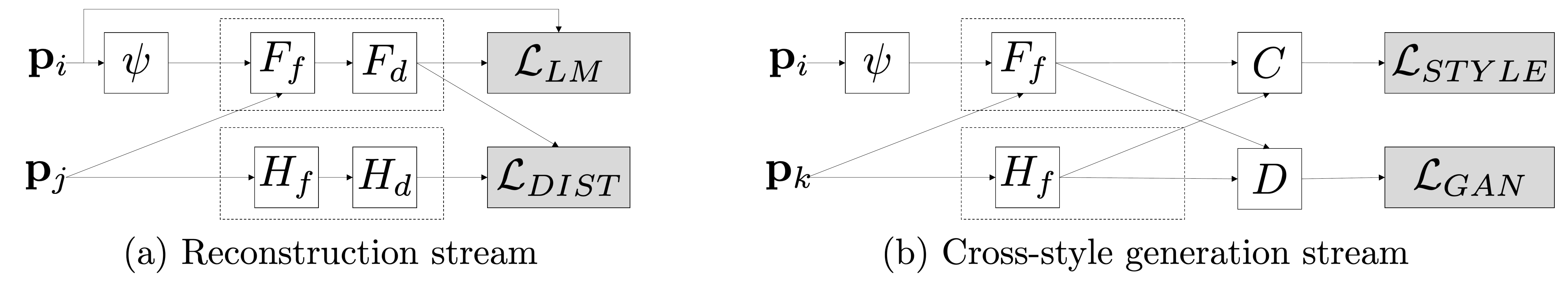}
    \vspace{-3mm}
    \caption{\small We employ two data streams to train our framework. While $\rvp_i$ and $\rvp_j$ have the same style, $\rvp_i$ and $\rvp_k$ do not. (a) The reconstruction stream is trained using the language modeling loss $\gL_{LM}$ and the distillation loss $\gL_{DIST}$. (b) The cross-style generation stream is trained using the style loss $\gL_{STYLE}$ and the GAN loss $\gL_{GAN}$. Note that we decompose each network into a feature extractor and an embedding layer.}
\vspace{-3mm}
    \label{fig:framework}
\end{figure}

We propose an objective function consisting of four carefully designed loss terms for training the proposed framework using the above two data streams. The objective function is given by 
\begin{equation}
\gL = \gL_{LM} + \lambda_{DIST}\gL_{DIST} + \lambda_{STYLE}\gL_{STYLE} + \lambda_{GAN}\gL_{GAN},
\end{equation}
where $\gL_{LM}$ is the language modeling loss,  $\gL_{DIST}$ the distillation loss, $\gL_{STYLE}$ is a style comparison loss, and $\gL_{GAN}$ is the latent GAN loss. The scalars $\lambda_{DIST}$, $\lambda_{STYLE}$, and $\lambda_{GAN}$ are the hyper-parameters controlling relative importance of the terms. The values for these hyperparameters and the method for determining their values are discussed in Appendix \ref{appendix:hyper_parameters}. We visualizes training with the proposed objective function using the two data streams in Fig.~\ref{fig:framework}.

{\bf Language modeling loss $\gL_{LM}$} formulates the probability distribution of a paragraph $\rvp$ as the product of the conditional probability of each token $w_{t}$ given the previous tokens $\{w_{t}\}_{t}^{T-1}$ as shown in~(\ref{equ:LM}). We use $\gL_{LM}$ to supervise the training of the data reconstruction stream. It is given by 
\begin{equation}
\gL_{LM} = E_{(\rvp_i, \rvp_j)\sim \texttt{RS}} \big{[}\frac{-1}{T}\sum_{t}^{T}log(\frac{e^{F(w_{t}|w_{1:t-1}, \rvp_j)}}{\sum_{v}^{V}e^{F(w_{v}|w_{1:t-1}, \rvp_j)}})\big{]},
\end{equation}
where $(\rvp_i, \rvp_j)\sim \texttt{RS}$ denotes that $\rvp_i$ and $\rvp_j$ are from the reconstruction stream. The variable $T$ is the total number of tokens in $\rvp_i$ and $V$ is the size of the vocabulary.

{\bf Distillation loss}. We use $\gL_{DIST}$ to regularize the learning as processing the data reconstruction steam. We pretrain a \gpttwo model using our dataset $\sD$ and use it as our distillation target. We denote the pretrained \gpttwo model as $H$. (Note that $H$ does not have the desired style control capability.) By jointly optimizing $\gL_{LM}$ and $\gL_{DIST}$, we train $F$ to generate fluent texts (by minimizing $\gL_{LM}$) as well as behave similarly to $H$ (by minimizing $\gL_{DIST}$). The distillation loss is calculated by minimizing the mutual information between output distributions of $F$ and $H$, which is given by    
\begin{equation}
	\gL_{DIST} = E_{(\rvp_i, \rvp_j)\sim \texttt{RS}} \big{[} \frac{-1}{T}\sum_{t}^{T}\sum_{v}^{|\nu|}\frac{e^{H(w_{t=v}|w_{1:t-1})}}{\sum_{v}^{|\nu|}e^{H(w_{v}|w_{1:t-1})}}log(\frac{e^{F(w_{t}|w_{1:t-1}, \rvp_j)}}{\sum_{v}^{|\nu|}e^{F(w_{v}|w_{1:t-1}, \rvp_j)}})  \big{]}.
\end{equation}
We note that the distillation loss has been used in various tasks including model compression, transfer learning, life-long learning, etc (\cite{hinton2015distilling, kim2016sequence, liu2019multi, mirzadeh2019improved, liu2019improving, hou2018lifelong}). In this paper, we extend its use to the style example-guided language generative model training task. 
	
{\bf Style loss $\gL_{STYLE}$} helps ensure  the output from the cross-style generation stream has the same style as the input reference. A pretrained style comparator $C$ is used for computing the loss. The comparator takes two paragraphs as input and is trained to output 1 when the two paragraphs have the same style and 0 otherwise. We use $\sD$ for pretraining $C$ since it contains style labels for each document. We pretrain $C$ using the binary cross entropy loss. The comparator $C$ is highly accurate. It achieves a classification accuracy of {87.8}\% to {98.8}\% in our held-out validation sets. After pretaining, we fix $C$ and use it to train $F$. The style loss $\gL_{STYLE}$ is then given by 
\begin{equation}
\gL_{C} = E_{(\rvp_i, \rvp_k)\sim \texttt{CS}} \Big{[}-\log \Big{(} C\big{(} 
H_f (\rvp_k), F_f(\psi(\rvp_i), \rvp_k)
\big{)}\Big{)}\Big{]}
\end{equation}
where $(\rvp_i, \rvp_k)\sim \texttt{CS}$ denotes the pair is sampled from the cross-style generation stream.

Here, we would like to make two remarks. First, since $C$ takes the latent feature from $F_f$ as input, we avoid the non-differentiability of the text decoding mechanism and can directly train $F_f$. Second, despite that $C$ is pretrained using feature extracted from $H_f$, we use the feature extracted from $F_f$ as input. We can perform this operation not only because these two features have the same dimension but also because we enforce them to have a similar distribution via optimizing the GAN loss, discussed below.

{\bf GAN loss $\gL_{GAN}$} is used to match the distributions of the features generated by $F_f$ and those generated by $H_f$, respectively, as processing the cross-style generation stream. We use a latent GAN formulation where we train a GAN discriminator $D$ to differentiate features extracted from $F_f$ to $H_f$. The GAN loss is given by
\begin{equation}
E_{\rvp_i, \rvp_k}\Big{[}-\log\big{(} D( H_f(\rvp_k) \big{)} - \log\big{(} 1 - D(F_f(f(\rvp_i), \rvp_k))\big{)}\Big{]}.
\end{equation}
We realize the discriminator $D$ using a \gpttwo-based transformer network.

\subsection{Generator Design}

\begin{figure}[!t]
    \centering
    \includegraphics[width=.99\linewidth]{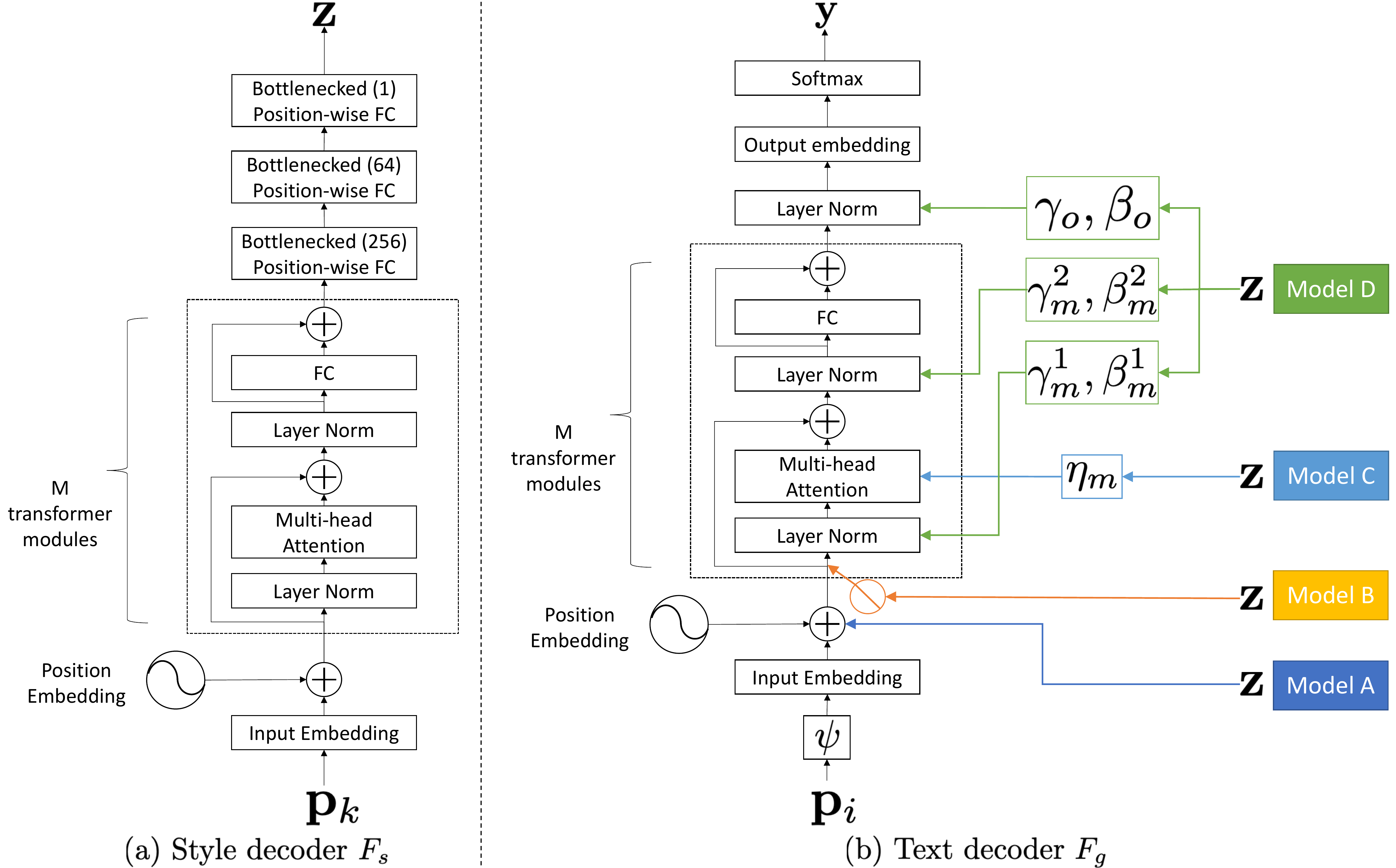}
    \vspace{-2mm}
    \caption{\small (a) Style encoder $F_s$. (b) Text decoder $F_g$. We consider 4 different ways of injecting the style code $\rvz$ into $F_g$ termed Model A, B, C, and D. Checkout main texts for more details.}
\vspace{-3mm}
    \label{fig:network}
\end{figure}

We realize the style encoder $F_s$ using a \gpttwo-based transformer identical to $H_f$. After extracting a representation $\rvz_{t}$ for each token $t$ in $\rvs$, we utilize a 3-layer position-wise fully-connected network to obtain the final style code $\rvz$ as illustrated in Fig.~\ref{fig:network}. The text decoder $F_g$ is also a \gpttwo-based transformer identical to $H$. We initialize the weights in $F_s$ and $F_g$ using the weights in the pretrained $H$. Next, we compare four different ways of injecting outputs from $F_s$ into $F_g$, which represent different inductive biases and result in difference performances.

{\bf Model A: style code as a bias to the input}. In this model, the style code $\rvz=F_s(\rvs)$ is directly summed up with the token-embedding and position embedding before inputting to the first transformer module in $F_g$. In other words, the input to the first transformer module in $F_g$ is $\rve_{t}^{w} + \rve_{t}^{p} + \rvz$ where $\rve_{t}^{w}$ denotes as the $t$th word embedding, and $\rve_{t}^{p}$ denotes as the $t$th position embedding.

{\bf Model B: style code as a summarization token}. In this model, the computed style code $\rvz=F_s(\rvs)$ is treated as a special token that is inserted to the beginning of the input sequence and is directed fed in the first transformer module in $F_g$. That is the input sequence length becomes $T+1$. This design is motivated by the traditional sequence-to-sequence modeling techniques~\citepara{chung2014empirical,cho2014learning, sutskever2014sequence,bahdanau2014neural,vinyals2015neural}.

{\bf Model C: style-aware self-attention}. In this model, we input $\rvz$ into each self-attention layer in $F_{g}$ to influence its computation given by
$Softmax(\frac{\rvq_m  \rvk_{m-1}^{T}}{\sqrt{B}})\rvv_{m-1}$ 
where $\rvq_m = \eta_m(\rvz)$ which $\eta_m$ denotes an affine transformation, $\rvk_{m-1}$ and $\rvv_{m-1}$ denotes the key and value embeddings from the $(m-1)$th hidden layer, and $B$ denotes the hidden dimension.

{\bf Model D: adaptive layer normalization}. Inspired by the recent success in image generation tasks~\citepara{park2019semantic, karras2019style}, we utilize the style code to modulate the hidden representations within the text decoder via normalization layers. Specifically, we replace the scale and bias parameters in the affine transformation step of the layer normalization~\citepara{ba2016layer} with a style code determined scale and bias. That is 
\begin{equation}
\gamma_{m,c}^{a}(\rvz) \frac{h_{m,c, t}^{a} - \mu_{m,t}^{a}}{\sigma_{m,t}^{a}} + \beta_{m,c}^{a}(\rvz),
\end{equation}
where $h_{m,c, t}^{a}$ denotes the $c$th hidden representation of the $t$th token at the $m$th transformer layer. We note $a=\{1,2\}$ since there are two layer normalization layers in each transformer in our implementation. The mean and deviation $\mu_{m,t}^{a}$ and $\sigma_{m,t}^{a}$ are computed across the channel dimension.

We illustrate how these models inject $\rvz$ to $F_g$ in Fig.~\ref{fig:network}. In Section~\ref{sec:expr}, we compare the performance of these variants and show that Model D achieves the best style generation performance.
\section{Experiments}\label{sec:expr}

{\bf Implementation}. We set the latent dimension $B$ to $768$, number of attention-heads to $16$, number of transformer layers $M$ to $16$, number of tokens in a paragraph $T$ to $512$, and the vocabulary size $V$ to $50257$ using BPE-encoding \citepara{sennrich2015neural} vocabulary from \cite{radford2019language} throughout out all the models and experiments. We use a pretrained \gpttwo model $H$ and a style comparator $C$ in our framework. The training details of these two models are given in Appendix~\ref{appendix:pretraining}. All of the experiments are conducted using an NVIDIA DGX1 machine.

{\bf Datasets}. We compare competing methods using two newly composed datasets based on~\citepara{zhu2015aligning, zellers2019neuralfakenews, wikixml2015, see2017get}.

\texttt{3-Style}. The dataset consists of documents from the RealNews dataset~\citepara{zellers2019neuralfakenews}, the BookCorpus dataset~\citepara{zhu2015aligning}, and the Reviews dataset~\citepara{yelp2019, mcauley2013amateurs, maas2011learning, hotel2017, 515khotel2017}. The 3 styles are news, book, and review. In detail, the news set has $33M$ documents and $113B$ words, the books set has  $50K$ documents and $7.2B$ words, and the review set has $4.8M$ documents and $5.4B$ words after cleaning. The total dataset has $37.85M$ documents and $125.6B$ words. We hold out $3.78M$ documents as the validation set and $6K$ documents as the testing set.

\texttt{21-Style}. We build a dataset that contains 21 text styles. We first classify the documents in RealNews into $9$ styles, including Sciences, Sport, Politics, Business, Technology, Entertainment, Opinion, Life, and News. Then, we divide the documents in BookCorpus into 8 different styles, which are Romance, Fantasy, Sciencefiction, Children’sbooks, Thriller, Adventure, Poetry, and Plays. We split the documents into multiple small documents by extracting the dialogues except for the Poetry and Plays. We divide the Review dataset into $3$ styles, namely Yelp, Hotel, and Movie. Finally, we crawl $0.77M$ lyrics from \url{http://www.azlyrics.com/}. The total dataset has $35.5M$ documents. We hold out $3.55M$ documents as the validation set and $21K$ documents as the testing set.

{\bf Auto-evaluation metrics}. We evaluate different models using fluency score, style score, style diversity score, and content novelty score. The fluency score measures whether the output paragraph reads like a human-written one. The style score checks whether the output text carries the target style. Our framework supports multimodal outputs~\citepara{huang2018multimodal}. For the same input context but different reference examples of the same style, our framework should produce different output texts but all with the same style. To measure how different these outputs are, we use the style diversity score. Finally, the content novelty score is used to measure the difference between the output and the reference example. A model that directly duplicates the reference to the output is undesirable. The details of these automatic evaluation metrics are available in Appendix~\ref{appendix:auto}.    

{\bf Human study settings}. We use the Amazon Mechanical Turk (AMT) platform for user studies. We conduct two studies where one evaluates fluency of the generated paragraphs while the other verifies the style correctness. For the fluency study, we present a human-written text and a machine-generated text in random order and ask the worker to choose which one is written by a human. For this metric, the closer the preference score to $50\%$, the better the performance. 

For the style study, we perform two tests. In one test, we present a worker a generated paragraph that supposes to be in the target style. We also give the worker two human-written reference paragraphs where one is with the target style while the other is not. We then ask the worker to choose which reference paragraph has a style more similar to the generated one. In the other test, we again present a worker a generated paragraph but this time with the style categorical labels to choose from instead of the reference paragraphs. We compute the frequency that the worker selects the right style. The higher the score, the better the performance. More details are in Appendix \ref{appendix:human_evaluation}.

{\bf Strong baselines}. We compare our framework to three strong baselines, namely the general \gpttwo model (\texttt{G-\gpttwo}), a baseline consists of multiple style-specialized \gpttwo models (\texttt{S-\gpttwo}), and the style-code encoding (\texttt{SC}) method based on the description in~\cite{Keskar2019languagegeneration}. \texttt{G-\gpttwo} is trained on the entire dataset using $\mathcal{L_{LM}}$. It does not allow style control but can generate fluent texts. In \texttt{S-\gpttwo}, we train a \gpttwo model per style. As training a \gpttwo model is costly, we only use this baseline for the \texttt{3-Style} dataset evaluation. In \texttt{SC}, an one-hot encoding of the style class label is used as a special token for style-controllable paragraph generation. Unlike the proposed method that extracts the style code from the input paragraph, \texttt{SC} input the style label. The rest of the model is similar to our Model B without the style encoder.

\subsection{Results}

\begin{figure}[!t]
\begin{subfigure}{.5\textwidth}
  \centering
  \includegraphics[width=\linewidth]{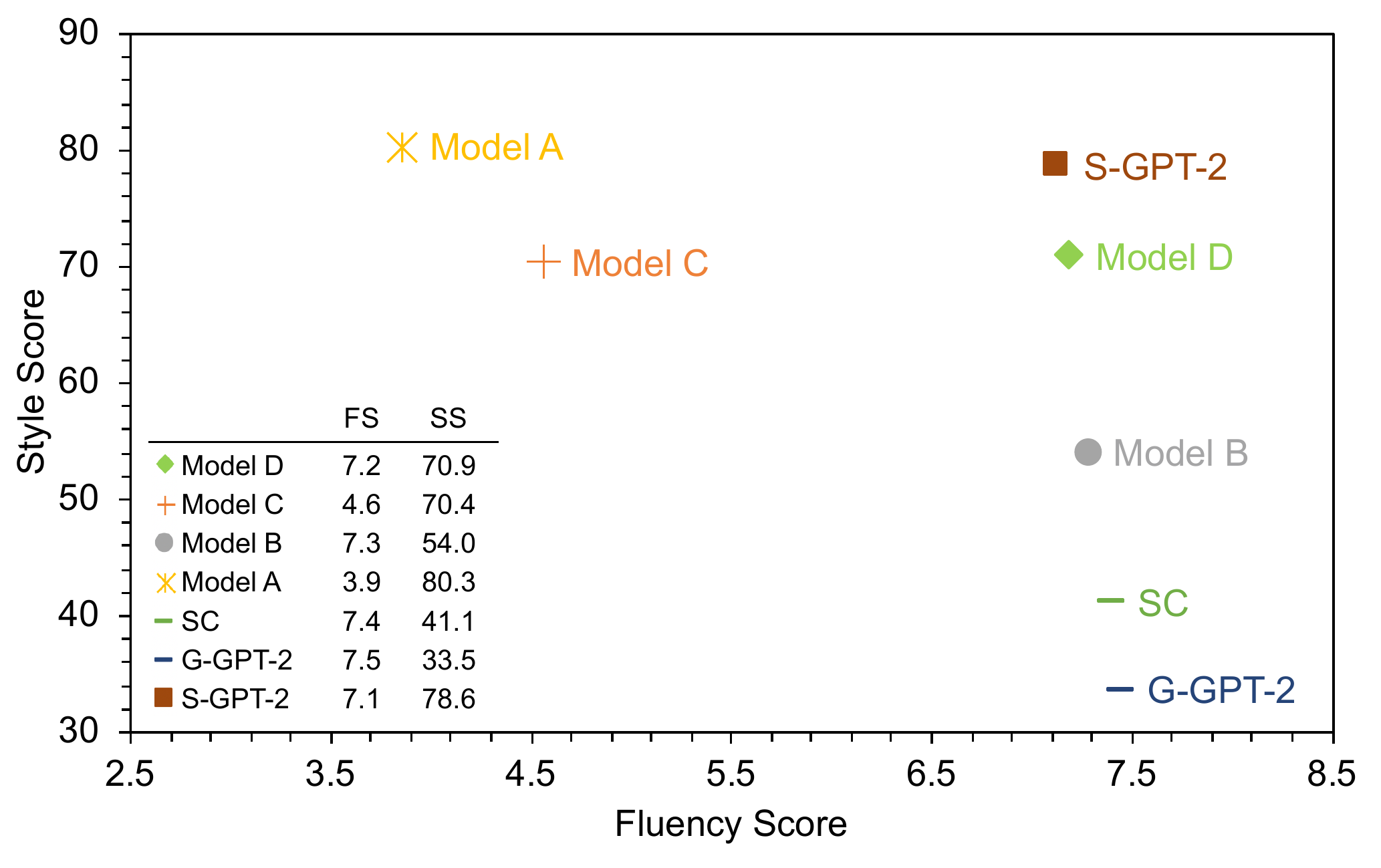}\vspace{-3mm}
  \caption{\small Fluency vs. style on the 3-Style dataset.}
  \label{fig:sfig1}
\end{subfigure}%
\begin{subfigure}{.5\textwidth}
  \centering
  \includegraphics[width=\linewidth]{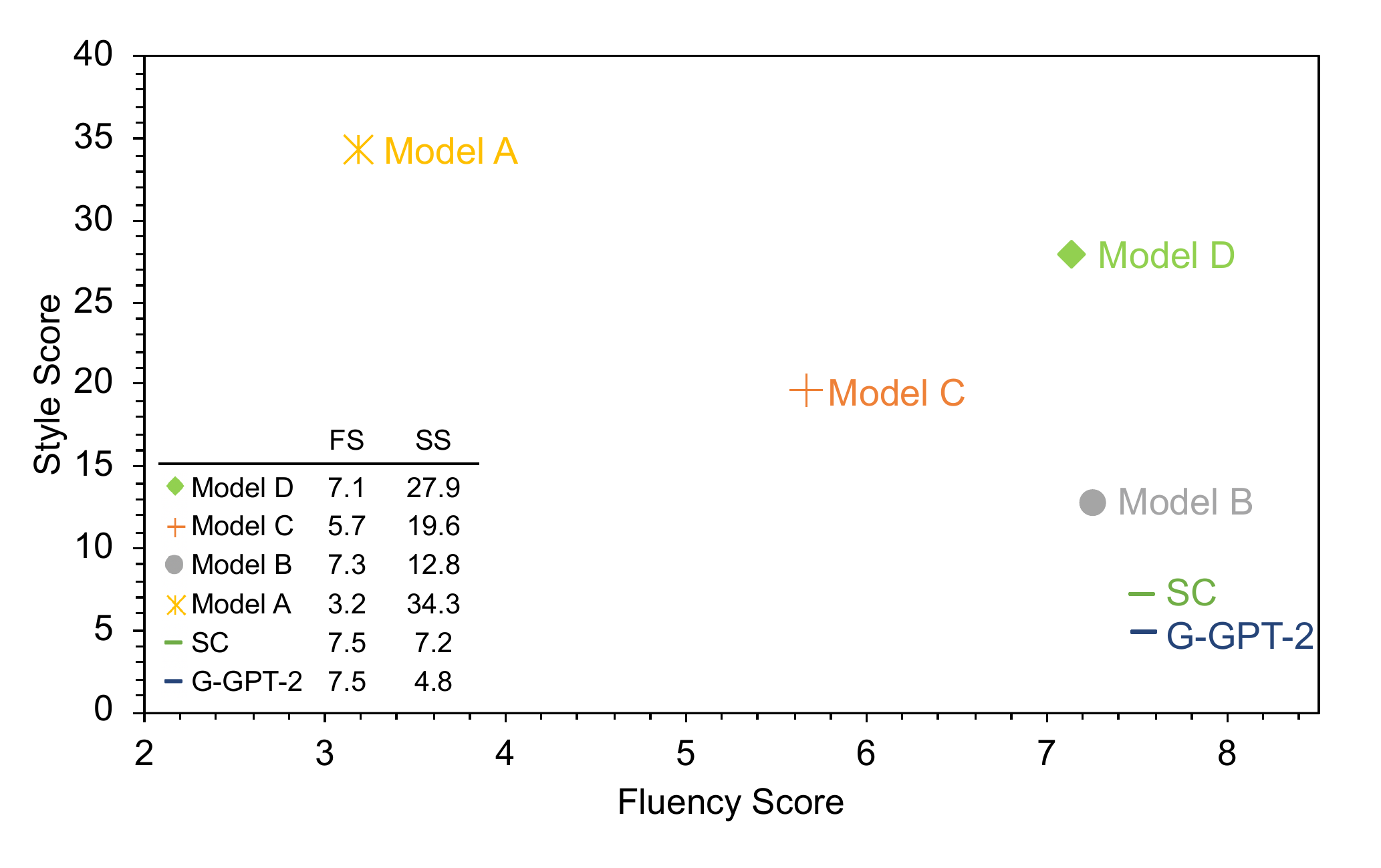}\vspace{-3mm}
  \caption{\small Fluency vs. style on the 21-Style dataset.}
  \label{fig:sfig2}
\end{subfigure}
\vspace{-3mm}
\caption{\small Fluency and style scores achieved by the competing models on the \texttt{3-Style} and \texttt{21-Style} datasets.}
\label{fig:fluency_vs_style}
\vspace{-2mm}
\end{figure}

In Fig.~\ref{fig:fluency_vs_style}, we plot the fluency and style scores achieved by our models as well as those by the baselines on the $3$-Style and $21$-Style datasets. The closer the model to the top-right corner, the more superior the model is. From the figure, we found that among our models, Model D performs the best. As expected, \texttt{G-\gpttwo} achieves the best fluency score. However, since it does not support style control, it has a poor style score. On the other hand, \texttt{S-\gpttwo} achieves good fluency and style scores for the \texttt{3-Style} dataset. This is understandable as it utilizes a \gpttwo model for each style. However, such an approach does not scale well as \gpttwo training is expensive. We also found that \texttt{SC} does not achieve good style score and is inferior to our models. We suspect this is because the one-hot style class code is largely ignored during inference. Since Model D performs the best in our framework, for the rest of the paper, we use it as our representative model for performance comparison as well as ablation study.

In Tab.~\ref{tab:taser}, we show several generation results from our Model D. We find that the output texts are fluent and respect the styles of the references. More output examples are available in Appendix~\ref{appendix:results}. 

In Tab.~\ref{tab:diversity}, we show the style diversity scores achieved by our models. We found that all of our 4 models can generate diverse styled paragraphs conditioning on the same context and different reference examples with the same style.

\begin{table}[!t]
    \centering
\resizebox{0.60\linewidth}{!}{\mbox{%
		{\tabcolsep=10pt\def\arraystretch{1}
	\begin{tabular}{|c || c | c | c | c |} 
	    \hline
	    & Model A  & Model B & Model C & Model D  \\ 
	    \hline
	    \multicolumn{1}{|c||}{\texttt{3-Style}}& \textbf{11.53} & 11.61 & 10.58 & 11.13   \\ \hline	    
	    \multicolumn{1}{|c||}{\texttt{21-Style}}& 10.17 & 10.09 & \textbf{11.32} & 10.52 \\ \hline
	\end{tabular}
}}}\vspace{-2mm}
	\caption{\small Style diversity scores achieved by the computing methods. We note the lower bound and upper bound for the style diversity scores are  4.52 and 15.67, respectively.}
	\label{tab:diversity}
	\vspace{-4mm}
\end{table}
\begin{table}[!t]
    \centering
    \begin{tabular}{cc}
\resizebox{0.43\linewidth}{!}{\mbox{{\tabcolsep=10pt\def\arraystretch{1}
	\begin{tabular}{|c || c | c || c |}
	    \hline
	    (\%) & Model D & \texttt{SC} & Random\\ 
	    \hline
	    \multicolumn{1}{|c||}{\texttt{3-Style}}&  56 & \textbf{54} & 50 \\ \hline	    
	    \multicolumn{1}{|c||}{\texttt{21-Style}}&  \textbf{57} & 63 & 50 \\ \hline
	\end{tabular}
}}} &
\resizebox{0.53\linewidth}{!}{\mbox{%
		{\tabcolsep=10pt\def\arraystretch{1}
	\begin{tabular}{c || c | c | c |}
	\hline
	\multicolumn{1}{|c||}{(\%)}& Model D & \texttt{SC} & Random\\ 
	\hline
	\multicolumn{1}{|c||}{\texttt{3-Style} by reference}&  \textbf{56} & 52 & 50 \\ \hline
	\multicolumn{1}{|c||}{\texttt{3-Style} by category}&  \textbf{65} & 54 & 50 \\ \hline	    
	\multicolumn{1}{|c||}{\texttt{21-Style} by reference}& \textbf{66} & 49 & 50 \\ \hline
	\multicolumn{1}{|c||}{\texttt{21-Style} by category}& \textbf{69} & 50 & 50 \\ \hline
\end{tabular}
}}}
\end{tabular}
\vspace{-2mm}
	\caption{\small (Left): Human study results on fluency. (Right): Human study results on style control. Random denotes the accuracy for random guess. Model D performs favorably over the baseline \texttt{SC}.}
	\label{tab:amt}\vspace{-4mm}
\end{table}
\begin{table}[!t]
\centering\resizebox{0.95\linewidth}{!}{\mbox{{\tabcolsep=10pt\def\arraystretch{1}
	\begin{tabular}{| c || c | c | c | c |}
	    \cline{1-5}
	    Model D & Fluency Score  &  Style Score  & Style Diversity Score & Content Novelty Score \\ 
	    \hline
	    \multicolumn{1}{|c||}{\st{ $\mathcal{L}_{DIST}$}}& 7.32 & 51.16 & 11.40 & 24.01\\ \hline
	    \multicolumn{1}{|c||}{ \st{$\mathcal{L}_{STYLE}$}} & 7.35 & 5.40 & 9.22 & 29.27\\ \hline
	    \multicolumn{1}{|c||}{ \st{$\mathcal{L}_{GAN}$}} & 6.85 & 28.67 & 10.35 & 26.77 \\ \hline
	    \multicolumn{1}{|c||}{All} & 7.14 & 27.90 & 10.52 & 25.85 \\ \hline
	\end{tabular}
}}} \vspace{-2mm}
	\caption{\small Ablation study on the various loss terms in the proposed objective function.}
	\label{tab:ablation_study_21}\vspace{-2mm}
\end{table}

{\bf Human evaluation}. In Tab.~\ref{tab:amt}, we report user study results on fluency and style control. We found that our model achieves great fluency on both of the datasets. Compared to \texttt{SC}, our model performs better in controlling the style in the output texts.

{\bf Ablation study}. We conduct an ablation study on the loss terms in the proposed objective function and report the results in Tab.~ \ref{tab:ablation_study_21} using the $21$-Style dataset. The results show that each term is important. Removing $\mathcal{L}_{DIST}$ leads to a degraded content novelty score. Removing $\mathcal{L}_{STYLE}$ leads to a degraded style score, thought an improved fluency score and a content novelty score. Removing $\mathcal{L}_{GAN}$ leads to both degraded fluency and style diversity scores.

\section{Conclusion}

We presented a language generative framework for style example-guided paragraph generation. To the best of our knowledge, we were the first to achieve such style-controllability on paragraph generation. We attributed the success to our carefully designed learning objective function, the generator network, and the newly composed large-scale dataset consisting of documents of various text styles.

\bibliography{iclr2020_conference}
\bibliographystyle{iclr2020_conference}

\clearpage
\vfill
\appendix

\section{Hyper-parameters Tunning}\label{appendix:hyper_parameters}

We tune the hyper-parameters on a pilot-test dataset. This pilot-test dataset has $3K$ training examples and $400$ hold-out examples. We perform a grid search on log-scale. We utilize the Frechet Embedding Distance~\citepara{semeniuta2018accurate} to select best hyper-parameters. For $\lambda_{DIST}$, $\lambda_{STYLE}$, and $\lambda_{GAN}$, we try $\{1.0, 0.1, 0.1\}$, $\{0.1, 0.1, 0.1\}$, and $\{1.0, 0.01, 0.01\}$. For \textit{betas} used in \textit{Adam} Optimizer, we try $\{0.9, 0.999\}$ and $\{0.0, 0.9\}$. For weight decay, we try $0.01$ and $0.0$. For the initial learning rate, we try $0.00025$ and $0.000025$. Eventually, we use $0.00025$ as initial learning rate and $\{1.0, 0.01, 0.01\}$ for $\lambda_{DIST}$, $\lambda_{STYLE}$, as well as $\lambda_{GAN}$ for all models except for the Model B. We use $\{1.0, 0.01, 0.1\}$ for the Model B instead.

\section{Pretraining of $H$ and $C$}\label{appendix:pretraining}

{\bf Pretraining of the GPT-2 model $H$}. We pretrain $H$ on our collected dataset $\sD$ from scratch. We use Adam optimizer with $0.00015$ initial learning rate, $\beta_{1}$ and $\beta_{2}$ are set by $\{0.9, 0.999\}$, cosine learning rate decay style, and $0.01$ weight decay. The batch size is set to $512$. The total training iterations is $320K$ which takes $2$ weeks.

{\bf Pretraining of the style comparator $C$}. We pretrain the Style Comparator $C$ using $256$ and $512$ batch size, respectively. The initial learning rate is $0.00015$ with $0.01$ weight decay and cosine learning rate decay style. The optimizer is also Adam with $\beta_{1}=0.9$ and $\beta_{2}=0.999$. Since the training converges very quickly, we do early stopping if the accuracy on validation set plateaus. Eventually, we use the checkpoint at $28K$ and $50K$ iterations to train on the $3$-style and $21$-style datasets, respectively. The accuracy on hold-out validation set are $98.8\%$ and $87.8\%$ on $3$-styles and $21$-styles datasets, respectively.

\section{Auto-evaluation Metrics}\label{appendix:auto}

{\bf Fluency score}. To ensure the generated paragraph is fluent and coherent, we utilize a pretrained \gpttwo model to measure the perplexity of the generated paragraph. We compute the generation likelihood over each token using the model and treat the generated paragraph $\rvy_{0:T-1}$ as inputs and $\rvy_{1:T}$ as labels. Given the input text, the pretrained GPT-2 returns the probability distribution of next token over the vocabulary. Then, we measure the perplexity by this probability distribution and label. Since our dictionary size is $50257$, the random guess of the next token would result in $50257$ perplexity ($\mathcal{L}_{LM} = log(50257) \approx 10.83$). Thus, we set $10.83$ as an upper bound and define the fluency score of the generated paragraph as $10.83-log(perplexity)$. In this sense, a higher fluency score means lower perplexity.

{\bf Style score}. We train 3/21 binary style classifiers (since we have 3/21 different styles in the \texttt{3-Style}/\texttt{21-Style} dataset) by finetuning a \gpttwo network to automatically evaluate whether the generated text carries the style of a target class. These 3/21 classifiers achieve average classification accuracies of $99.1\%$/$96.3\%$. During the testing phase, for a target style, if the corresponding style classifier correctly predicts 1 for the generated paragraph computed by a model, we count it as a successful trial. We compute the success rate over the test set and use the result as the style score for the model.

{\bf Style diversity score}. We adopt the LPIPS distance~\citepara{zhang2018unreasonable} to measure the diversity of the generation outputs conditioning on the same context. To implement this metric, we first extract the feature representation from each token in a generated paragraph by a pretrained \gpttwo model. We compute the mean representation of the tokens in a paragraph as the paragraph-level representation. Then, we measure the $L_{2}$ distance between two paragraph-level representations of two different paragraphs generated using the same context but two different references written in the same style. In this sense, a larger distance value implies the styles of the two generated paragraphs are more different. 

To get an idea of the range of this metric, we compute an upper bound and a lower bound. We consider two paragraphs from two documents of different styles should have a high style diversity score. We hence sample $1000$ paragraphs from each style and use the pretrained \gpttwo model to extract deep features. After taking average over the token-dimension, we obtain $1000$ by $768$ representation for each style. Then, we compute the $\mathcal{L}_{2}$ distance between of these matrices divided by 1000. This gives us a matrix of size $21\times21$ measuring the pairwise distance between two styles. We use the largest value in this matrix as our upper bound, which is $15.67$.

For the lower bound, since two different paragraphs from the same document should have a low style diversity score, we use their scores for the lower bound computation. Specifically, we compute the average distance between two different paragraphs from the same document. We do this for each style and obtain 21 different values. We obtain the lower bound by taking average over these values, which is $4.52$.

{\bf Content novelty score}. To verify that our model is not simply duplicating the content from reference paragraph, we utilize LPIPS distance~\citepara{zhang2018unreasonable} to measure the difference between the generated paragraph and the input reference paragraph. We again use a pretrained \gpttwo model for extracting a feature representation for each token. To compute the distance between two paragraphs, we compute the bipartite matching cost between the tokens from the two paragraphs. Specifically, we first compute the $L2$ distances between any token representation in one paragraph to all the token representations in the other paragraph. We then compute the minimum cost assignment by solving a bipartite matching problem. In order to get an idea about the range of the content novelty score, we compute an upper bound and a lower bound using a similar approach as the one used for the style diversity score. We find the upper bound value is $31.46$ and the lower bound value is $15.13$.

\section{Human Evaluation}\label{appendix:human_evaluation}

\begin{figure}[h]
	\centering
	\includegraphics[width=\linewidth]{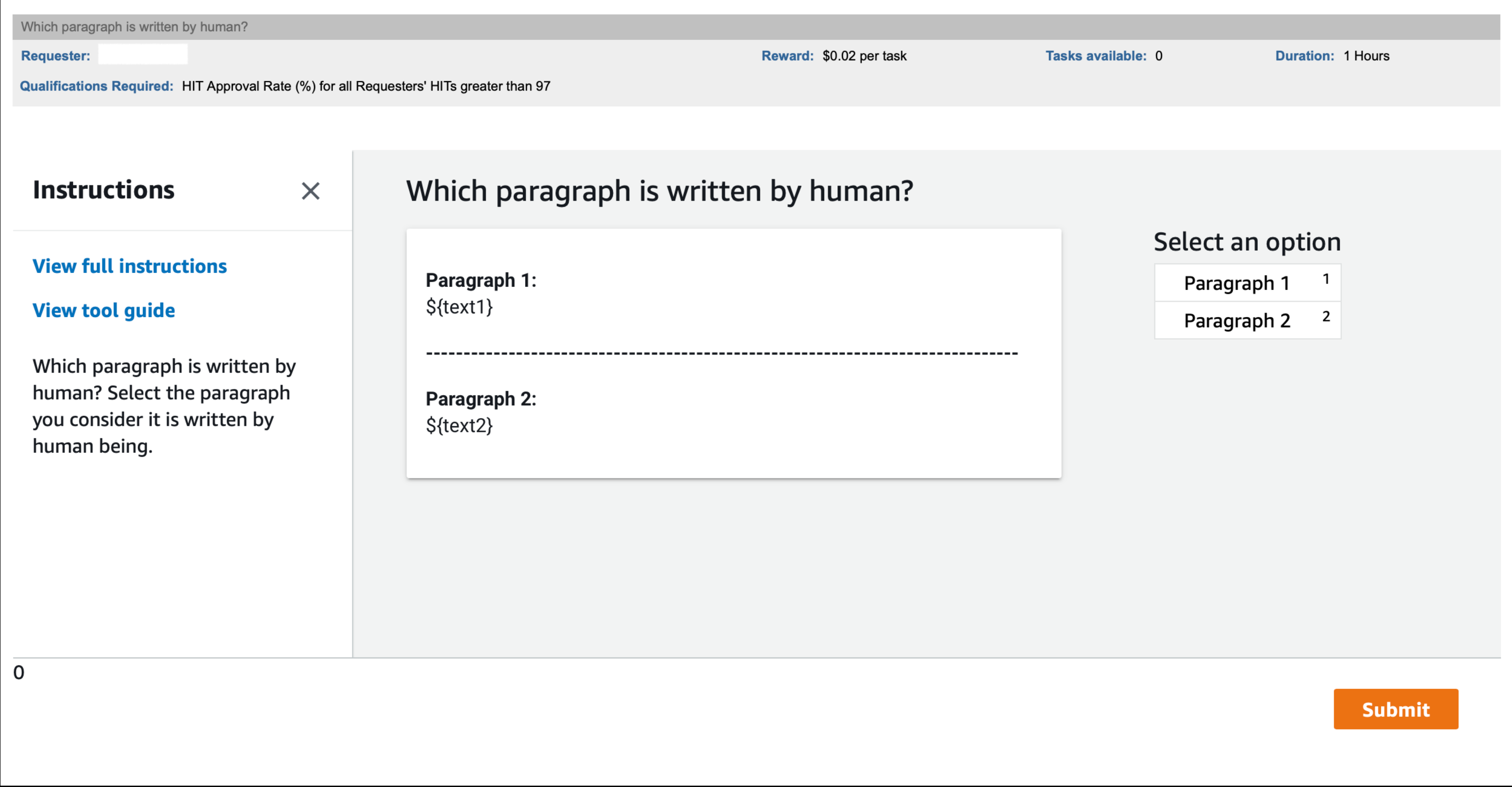}
	\vspace{-2mm}
	\caption{A typical example of the questionnaire for our A/B test on AMT for the fluency study.}
	\label{fig:amt_typical_example}
\end{figure}

To participate in our user study, a worker has to be awarded the Master Qualification by AMT and has at least $97\%$ life-long HIT approval rate. We generate $2$ paragraphs for a context with $2$ different reference paragraphs from the same style. For $3$-style dataset, we randomly sample $50$ examples from each style and do style generation for all $3$ styles. Thus, there are $900$ examples for each testing model. For $21$-style dataset, we randomly sample $42$ examples from testing set and do style generation on all $21$ styles. Thus, there are $1764$ examples for each testing model. There is a typical example for experiment of fluency human study in Fig. \ref{fig:amt_typical_example}.

\section{Additional Results}\label{appendix:results}
In Tab.~\ref{tab:app1},~\ref{tab:app2},~\ref{tab:app3},~\ref{tab:app4}, and~\ref{tab:app5}, we present additional example outputs of the proposed framework. The results demonstrate that the proposed model is fully capable of generating fluent output texts respecting the style of the input reference examples.

\begin{table}[h]
\begin{adjustbox}{width=\columnwidth,center}
	\centering
	\begin{tabular}{| l | l |}
	    \cline{1-2}
	    \begin{tabular}{U}
	    {\bf Context -- Poetry Style}\\
	    If anything, I'm a petty thief in a world of forgetting, a blade in an elm, 
	    a bottled note in the sea of Victory Boulevard at dusk.
	    Friends, family, gallows saints \& ghosts, rig a swing at the edge of the shore
	    when I'm gone for the birds to perch \& the bats to glide through,
	    to disappear \& appear in the small places the moon cannot touch.
	    Thank you for the amorous \& discontented beauty you saved for me here,
	    though most days, my silence was a soldier's slogged hand
	    hiding a match. I love you. I leave you the match.\\ \\ \\ \\ \\ \\
	    \end{tabular}
	    &
	    \begin{tabular}{l}
	    \\{\bf Reference -- Lyrics Style (Senorita)}\\
	    I love it when you call me senorita\\ 
	    I wish I could pretend I didn't need ya\\
	    But every touch is ooh la la la\\
	    It's true, la la la\\
	    Ooh, I should be running\\
	    Ooh, you keep me coming for you\\
	    Land in Miami\\
	    The air was hot from summer rain\\
	    Sweat dripping off me\\
	    Before I even knew her name, la la la\\
	    It felt like ooh la la la\\
	    \\
	    \end{tabular}\\
	    \hline
	    \multicolumn{2}{|l|}{\begin{tabular}{l}
	    \\
	    {\bf Generated}\\
	    I turn your body in my lap\\
	    your purse, a Pen, of a beautiful people\\
	    and I love you the way you love me, look at you drop a hand\\
	    \\
	    You wake up your ad libs Cirospis, cetachi, cetachi, cetachi, cetachi, cetachi, massageas, cetachi, cetachi, massageas, London Cirospis\\
	    I love you inside, I love you at the foot of your bed,\\
	    I love you inside, I love you at the foot of your bed,\\
	    \\
	    Leave your big collar, you leave your small collar, you leave your little collar\\
	    \\
	    Your life is just a half hour, it's your dance courses\\
	    you take a high vase of [real] ciche\\
	    just like our flowers\\
	    \\
	    We all want it so much, we want it so much, everybody\\
	    Then kiss thing, girl go kiss it\\
	    \\
	    I set the girl on her arms\\
	    against the state where no one could give it\\
	    --love is reserved for us, romanceis reserved for us, romanceis reserved for us, loveis reserved for us, loveis reserved for us,\\
	    \\
	    And it began at the state where no one could give it\\
	    --love is reserved for us, romanceis reserved for us\\
	    \\
	    --you are not just dresses but you are wearing them all a million times\\
	    , you are your own beautiful girls' effects\\
	    They blend in faster than the others\\
	    You meet and love all things, you . . .\\
	    \\
	    I am a proud Greek, an ancient Greek virgin\\
	    And I, like you, in high places, love all things, sexy\\
	    But on your early morning sleep, samba but in high places, sexy\\
	    Vesey no Vesey feather savant\\
	    \\
	    \end{tabular}}\\
	    \hline\hline
	    
	    \begin{tabular}{U}
	    {\bf Context -- News Style}\\
	    CARL AZUZ, CNN 10 ANCHOR: Hi, I`m Carl Azuz and you`re watching CNN 10.
	    Ten minutes of news explained. 
	    There are a lot of fast moving political headlines from the United States right now.
	    On Wednesday the White House is expected to release the transcript of a phone call.
	    A conversation between the presidents of the United States and Ukraine. 
	    The circumstances surrounding that call and something allegedly
	    discussed on it were the reasons why on Tuesday Democrats from the U.S.
	    House of Representatives announced they were moving forward on an impeachment inquiry, an investigation\\
	    \end{tabular}
	    &
	    \begin{tabular}{l}
	    \\
	    {\bf Reference -- Lyrics Style (Baby Shark)}\\
	    Baby Shark doo doo, doo doo doo doo\\
        Baby Shark doo doo, doo doo doo doo\\
        Baby Shark doo doo, doo doo doo doo\\
        Baby Shark\\
        Mummy Shark doo doo, doo doo doo doo\\
        Mummy Shark doo doo, doo doo doo doo\\
        Mummy Shark doo doo, doo doo doo doo\\
        Mummy Shark\\
        \\
	    \end{tabular}\\
	    \hline
	    \multicolumn{2}{|l|}{\begin{tabular}{l}
	    \\
	    {\bf Generated}\\
	    of a real, if not hoaxed, race in the United States.\\

All right, hey can I have a snack, Bob Bush?\\

A little snack baby!\\

Measures how to speed up the Country children?\\

Programs be grateful, Goodwill be giving.\\

Messages like, what is this thing about 8 am in the Middle?\\

Benjamin Franklin? Go\\

Nintendo Wii, Mario\\

Nintendo DS, U\\

Maria Barbara, Nice Guy\\

Johnny Cash, Mickey Mouse 2, Magic Tony 3,
Superman, Backzapper, Yellowjack, Bambi, Pig\\

His sister Diva, Saaath, Lil Saaath\\

A sad, sad, sad\\

If we only do than, what are he doing with it ?\\

Let's move forward and make it better, we can officially\\

make it good good, especially now Delma 0, Cadillac .\\

MAHA YEAH!!! :DDD\\

La Dara No Art, Chicken and Beach\\

(JCreates La Dara No Art, Chicken and Beach)\\
\\
	    \end{tabular}}\\
	    \hline
	    
	\end{tabular}
	\end{adjustbox}
	\caption{Qualitative Results: to lyrics style.}
	\label{tab:app1}
\end{table}

\begin{table}[h]
\begin{adjustbox}{width=1\columnwidth,center}
	\centering
	\begin{tabular}{| l | l |}
	    \cline{1-2}
	    
	    \begin{tabular}{U}\\
	    {\bf Context -- Lyrics Style (USA National Song)}\\
Oh, say can you see,\\
By the dawn's early light,\\
What so proudly we hailed,\\
At the twilight's last gleaming?\\
Whose broad stripes and bright\\
\\
	    \end{tabular}
	    &
	    \begin{tabular}{U}
	    {\bf Reference -- Hotel Reviews Style}\\
	    We loved our stay at Koloa Landing Resort. Wish we could afford to stay longer than a weekend. Rooms are gorgeous with lots of amenities - kitchen, free WiFi, washer, dryer, A/C and more. Great pool with water slides. Holoholo Grill has great tasting burgers and Kalua Nachos.\\ \\ 
	    \end{tabular}\\
	    \hline
	    \multicolumn{2}{|l|}{\begin{tabular}{Q}\\
	    {\bf Generated}\\
	    Satellites shared our love?\\
South Side Estate.\\
This is a great place to sample attorneys, chefs, therapists, etc. Dr Kirk Abad's is amazing! \\
\\
My family and friends stay at Sunset at Crinke Plaza for pretty cheap food and/or great view.
We took our daughter there because I hate visiting Vegas.\\
\\
Overall, even this rating is disappointing. "Doubt" is not a justifiable call for self-indulgence. It's a Deveuulment of humans. It's an attempt to give it more than its due. But, at least, today, Hollywood isn't offering it the world. That's what movie fans do with their cliches.\\
\\
Unless this story imagines a level of desperation that negative things just cannot get.\\
\\
	    \end{tabular}}\\
	    \hline\hline
	    \begin{tabular}{U}
	    {\bf Context -- News Style}\\
	    September 18, 2019
\\
Today, CNN 10 is explaining a recent rise and fall in oil prices and how they're tied to an attack on Saudi Arabian oil facilities. Afterward, a pair of scientific stories highlight studies on sharks and rats. And we're jumping into a 10 out of 10 segment featuring the world's largest bounce house.\\ \\ \\ \\ \\
	    \end{tabular}
	    &
	    \begin{tabular}{U}\\
	    {\bf Reference -- Movie Reviews Style}\\
	    Trailer was better than the movie. To me this movie just never takes off. It's a shame because there is some fantastic imagery and a wonderful theme lying underneath but in the end it can not overcome its plot holes and failed delivery of the message its trying to speak. It truly believes it's deeper than it is, but left me unfulfilled. Kind of lazy/rushed writing in my opinion. Kind of just throws a bunch of mediocre stuff at you without any real substance and expects the audience to give it purpose and deeper meaning in there own heads. Which is fine to do but to me this movie wasn't worthy of that merit of thought. You keep waiting for it to evolve but it just never does. Honestly can't believe critics are putting this on such a pedestal giving it such rave reviews.\\
	    \\
	    \end{tabular}\\
	    \hline
	    \multicolumn{2}{|l|}{\begin{tabular}{Q}\\
	    {\bf Generated}\\
	    Because CNN 10 just isn't accurate and does no justice. Instead, it conveys the film far too much to it's due and serves to give it more credibility. It's light sci-fi instead of enough fuel for actors helping to bring this premium piece down to its current disgustingness–especially not because that story is so good and necessary but because we live in a world where people like that can succeed.\\
\\
Overall, even this rating is disappointing. "Doubt" is not a justifiable call for self-indulgence. It's a Deveuulment of humans. It's an attempt to give it more than its due. But, at least, today, Hollywood isn't offering it the world. That's what movie fans do with their cliches.\\
\\
Unless this story imagines a level of desperation that negative things just cannot get.\\
\\
	    \end{tabular}}\\
	    \hline\hline
	    \begin{tabular}{U}
	    {\bf Context -- Paper Style (Abstract of this paper)}\\
We introduce a language generative model framework for generating a styled paragraph based on a context sentence and a style reference example. The framework consists of a style encoder and a texts decoder. The style encoder extracts a style code from the reference example, and the text decoder generates texts based on\\ \\ \\ \\
	    \end{tabular}
	    &
	    \begin{tabular}{U}\\
	    {\bf Reference -- Paper Reviews Style}\\
	    This paper proposes a new unsupervised learning approach based on maximizing the mutual information between the input and the representation. The results are strong across several image datasets. Essentially all of the reviewer's concerns were directly addressed in revisions of the paper, including additional experiments. The only weakness is that only image datasets were experimented with; however, the image-based experiments and comparisons are extensive. The reviewers and I all agree that the paper should be accepted, and I think it should be considered for an oral presentation.\\
	    \\
	    \end{tabular}\\
	    \hline
	    \multicolumn{2}{|l|}{\begin{tabular}{Q}\\
	    {\bf Generated}\\
	    a z and, if the pupils offer enough color to illuminate an image, the ones not enough are marked to be retouched.\\
\\
The grating is an issue with the words where thoughts can be expressed thought. However, both concepts are related, as well as relationships in general.\\
\\
For example, hate being done by children means that all books on life end matter:\\
\\
positive-positive-negative expectations, creating a model in which the writers at the top of Google and the bottom will be able to deliver claims regarding better quality than everyone else. There are theories in the research literature explicitly regarding these effects, but the mechanisms flating in elementary tasks that if they are crudely understood are generally significant (albeit less elaborated yet). A further test is to also decide:\\
\\
What is presented in the order\\
\\
	    \end{tabular}}\\
	    \hline\hline
	    \begin{tabular}{U}
	    {\bf Context -- Wiki Style (Alan Turing Bio)}\\
During the Second World War, Turing worked for the Government Code and Cypher School (GC\&CS) at Bletchley Park, Britain's codebreaking centre that produced Ultra intelligence. For a time he led Hut 8, the section that was responsible for German naval cryptanalysis. Here, he devised a number of techniques for speeding the breaking of German ciphers, including improvements to the pre-war Polish bombe method, an electromech\\ \\ \\ \\ \\ \\ \\ \\ \\ \\ \\ \\ \\ \\ \\
	    \end{tabular}
	    &
	    \begin{tabular}{U}\\
	    {\bf Reference -- Yelp Reviews Style}\\
	    I heard about SomiSomi from my Socal and Norcal friends before I even knew what it was! Taiyaki, soft serve, dessert goodness galore.\\
\\
We happened to get a taste of SomiSomi during their first day of soft opening inside the Westfield Valley Fair mall. They're located by the Men's Macy's, but also in PRIME location by Ramen Nagi and the new Pokeatery that will be opening soon apparently. There weren't a lot of seating options once we got our orders, which was a bummer. There were a few couches and more smaller seats, but when it gets busy like that, seating is important. Onto the food:\\
\\
We were given a free custard taiyaki with an ah-boong order, which is the soft serve in a cup and topped with a taiyaki of your choice. IT WAS AMAZING. Such a great deal. Pro Tip: Each somisomi has different mixed flavors. So I chose the milk tea mixed with ube and the taiyaki with taro. THE TARO FLAVOR IS THE BOMB. I don't think I could try anything else inside the taiyaki. Also, the taiyaki is made to perfection and are puffy delicious eggy-ness. It reminds me of Taiwan's egg puffs. The custard taiyaki was very tasty--not too sweet and the perfect consistency! They allow any and all toppings, and I really loved their coconut flakes. They also allow either a topping of a macaron or strawberry slice. I can't wait to come back and get some more SomiSomi.\\
\\
	    \end{tabular}\\
	    \hline
	    \multicolumn{2}{|l|}{\begin{tabular}{Q}\\
	    {\bf Generated}\\
	    propellant that helps vaporise the CJU Terminator atomic bombs, and other advancements so that the so-called uiorat! working meter could be controlled if its represented on a metal meter — not such a big deal if you'reme like me, or as someone who tends to enjoy standing in front of the clock.\\
There were also other inspirations from his Bletchley Park, namely Nieman K. Amer says:
"The best bits of Toni Sun's Kaki at Burien are Mavi Huli's. Sanjan and Caputo Kim (expectedly my favorite) sold out this ONE time and even there was part of our meeting there that made me queasy."\\
Apple: The raison d'etre resting in a gotcha
which was all tin from Customer Service.\\
Pocket: A new candy bar, brie strips (lots to choose from)\\
The secret, new dessert.\\
I love Crunchycris and all of the other dungbibats that will be available. It will melt on your lips while reading a book like The Fast and The Furious or any OMG crayon/fancy-rock. I want Tartan with a tangy sesame, Ma Tom Hagegai Fresh and Maaaadure - Bietan-Rose. But no miso ice cream cake or puns.\\
Be aware: get your shop prepared. Definitely not an artisanal place.\\
\\
	    \end{tabular}}\\
	    \hline
	    
	\end{tabular}
	\end{adjustbox}
	\caption{Qualitative Results: to reviews style.}
	\label{tab:app2}
\end{table}
\begin{table}[h]
\begin{adjustbox}{width=1\columnwidth,center}
	\centering
	\begin{tabular}{| l | l |}
	    \cline{1-2}
	    \begin{tabular}{U}
	    {\bf Context -- Children Books Style (Snow White and the Seven Dwarfs)}\\
	    Once upon a time . . . in a great castle, a Prince's daughter grew up happy and contented, in spite of a jealous stepmother. She was very pretty, with blue eyes and long black hair. Her skin was delicate and fair, and so she was called Snow White. Everyone was quite sure she would become very beautiful. Though her stepmother was a wicked woman, she too was very beautiful, and the magic mirror told her\\ \\ \\ \\ \\
	    \end{tabular}
	    &
	    \begin{tabular}{U}\\
	    {\bf Reference -- Poetry Style}\\
	    A black vulture crushes my bones.\\
A white vulture in a photograph. \\
Houses open their mouths, swallow
flowers. Lightning doesn't choose. I brushed
the spruces by hand. A car has a tire, a farmhand
a whip, trash cans are falling. We have
two natures. The vulture circles the belfry
at Ptujska Gora. Gingerbread hearts, red
blood, something awaits! Dawn bangs into
the bricks of Brooklyn. I'm still asleep
in silvery milk. When the wind strokes the grass,\\
I'll be in front of the house. Crickets will burn the air.\\
The sound will be like in the hall where
there's plush lining in Christ's tomb.\\
\\
	    \end{tabular}\\
	    \hline
	    \multicolumn{2}{|l|}{\begin{tabular}{Q}\\
	    {\bf Generated}\\
wait for my goddaughter to faint! For my stepmother is the thought of
greenery, gardens and swyrels!\\
\\
Still having a conversation scraped off the edge of my own couch,\\
\\
I hear the Ming river cry over the hearth,\\
\\
A tide will strain,\\
\\
and so will flowers.\\
\\
Up me through her hair,\\
\\
More flowers,\\
\\
A thistle bleeds,\\
\\
a urine of vibrant sugar.\\
\\
A red half of my fangs open and reach my face,\\
\\
Several holes in my chest.\\
\\
And a piano\\
\\
	    \end{tabular}}\\
	    \hline\hline
	    
	    \begin{tabular}{U}
	    {\bf Context -- Instagram Style}\\
An amazing meal had at @thestratford, testing their newly opened restaurant. Buttermilk deep-fried chicken with a doughnut and salad. An absolute must. Thank you @thestratford \#thestratford \#london \#londonrestaurant \#londonrestaurants \#cuisine \#food \#foodporn \#foodphotography \#foody \#foodgasm \#foodstagram \#foodlover\\ \\ \\ \\ \\ \\
	    \end{tabular}
	    &
	    \begin{tabular}{U}\\
	    {\bf Reference -- Poetry Style}\\
	    A black vulture crushes my bones.\\
A white vulture in a photograph. \\
Houses open their mouths, swallow
flowers. Lightning doesn't choose. I brushed
the spruces by hand. A car has a tire, a farmhand
a whip, trash cans are falling. We have
two natures. The vulture circles the belfry
at Ptujska Gora. Gingerbread hearts, red
blood, something awaits! Dawn bangs into
the bricks of Brooklyn. I'm still asleep
in silvery milk. When the wind strokes the grass,\\
I'll be in front of the house. Crickets will burn the air.\\
The sound will be like in the hall where
there's plush lining in Christ's tomb.\\
\\
	    \end{tabular}\\
	    \hline
	    \multicolumn{2}{|l|}{\begin{tabular}{Q}\\
	    {\bf Generated}\\
shows in the building news feeds, adorned with neon signs and mason jars. Just how much of the thrikers are griting
compassion, you know thoughts, for weeks burning in the hemispheres. I call my guests 'the
french!'s, the eggs and cream, the eggs and chicken…\\
\\
The Kinnear growling embers of bees tingle inside each woman. "Beat a cusha a crow."\\
\\
She pauses, looking at Anna, doubt sitting heavy on her face. "You are warm
as a fly. For you, those who want to live you will be happy,\\
\\
	    \end{tabular}}\\
	    \hline\hline
	    \begin{tabular}{U}
	    {\bf Context -- Sport News Style}\\
	    Stating the obvious apparently costs \$50,000 in the NBA.\\
\\
According to Shams Charania of Stadium and The Athletic, the league fined the Milwaukee Bucks after general manager Jon Horst told fans at a town hall event the Bucks plan to offer Giannis Antetokounmpo a supermax extension when they're eligible to do so.\\
\\
Horst referenced how league rules prevent Milwaukee from formally discussing specifics regarding what it would\\ \\ \\ \\ \\ \\
	    \end{tabular}
	    &
	    \begin{tabular}{U}\\
	    {\bf Reference -- Poetry Style}\\
	    It is written somewhere in the sky—we are supposed to forget quickly. How? An elephant stands on a tiny pedestal. In a field under a tent. The ringmaster spectacular in his tails and topcoat, expert in redirection, points towards the acrobat. The acrobat astounds because he could fall.\\
 \\
The other day I told my friend the story in which you fell through the floorboards of an old barn. As I told the story, I recalled the manner in which you carefully transferred milk to a newborn calf. You beckoned to Babette singing come Bossy Boss come Boss.\\
 \\
Years since you carved onto your forearm that pig with wings but have you ever noticed that the word pigeon encompasses the word pig? Once, in the evening on Fifth or Third, I saw a pigeon drop into a vat of mustard. I watched its dark wings thrash. Then yield.\\
\\
	    \end{tabular}\\
	    \hline
	    \multicolumn{2}{|l|}{\begin{tabular}{Q}\\
	    {\bf Generated}\\
earrudgeed its way in May for a warrant to call.\\
\\
After the conditions were waived, Indiana Pacers head coach Luke Walton and Bucks forward Patrick Ewing came into the meeting to see what those rules meant.\\
\\
So what were the rules?\\
\\
At first, we thought between letting training and verbal communication come into play. Now, with the way you communicate, you just have to make some other rational request.\\
\\
There's a tall castle rising in the sky, and a great sweet tale conjured up right. I cover a chandelier in a gift box; over the ledge you can spot a diamond with feathers on it. Learn to try again.\\
\\
We can't afford to \\
\\
	    \end{tabular}}\\
	    
	    \hline
	    
	\end{tabular}
	\end{adjustbox}
	\caption{Qualitative Results: to poetry style.}
	\label{tab:app3}
\end{table}
\begin{table}[h]
\begin{adjustbox}{width=1\columnwidth,center}
	\centering
	\begin{tabular}{| l | l |}
	    \cline{1-2}
	    \begin{tabular}{U}
	    {\bf Context -- Random sentence}\\
	    I have a cute dog\\ \\ \\ \\ \\ \\ \\ \\ \\ \\ \\ \\ \\ \\ \\ \\ \\ \\
	    \end{tabular}
	    &
	    \begin{tabular}{U}\\
	    {\bf Reference -- Children Books Style (Little Red Cap)}\\
	    "Good day to you, Little Red Cap."\\
\\
"Thank you, wolf."\\
\\
"Where are you going so early, Little Red Cap?"\\
\\
"To grandmother's."\\
\\
"And what are you carrying under your apron?"\\
\\
"Grandmother is sick and weak, and I am taking her some cake and wine. We baked yesterday, and they should be good for her and give her strength."\\
\\
"Little Red Cap, just where does your grandmother live?"\\
\\
"Her house is good quarter hour from here in the woods, under the three large oak trees. There's a hedge of hazel bushes there. You must know the place," said Little Red Cap.\\
\\
	    \end{tabular}\\
	    \hline
	    \multicolumn{2}{|l|}{\begin{tabular}{Q}\\
	    {\bf Generated}\\
"Well, I found a small white dog worth the trouble."\\
"What species are my family?"\\
"I'm sure of something. She is a farm."\\
"It's been a while since your last Christmas story," said George.\\
"I give up and hang up my Christmas album."\\
"Good, how many stores on earth now is the size of Big Red?"\\
He didn't feel right.\\
"I suppose ten," he said\\
\\
	    \end{tabular}}\\
	    \hline\hline
	    
	    \begin{tabular}{U}
	    {\bf Context -- Yelp Style}\\
Somisomi is easily one of my go-to spots for dessert because there's no lactose in the ice cream, it's very instagrammable, and they have rotating flavors.\\
\\
My friend and I decided to visit this location a few days after it opened and surprisingly, there was a really short line. This location had 6 flavors (matcha, milk, oreo, ube, coffee, and milk tea). What\\ \\ \\ \\ \\ \\ \\ \\ \\ \\ \\ \\ \\ \\ \\ \\ \\ \\ \\ \\ \\ \\ \\ \\

	    \end{tabular}
	    &
	    \begin{tabular}{U}\\
	    {\bf Reference -- Children Books Style (Snow White and the Seven Dwarfs)}\\
	    Once upon a time . . . in a great castle, a Prince's daughter grew up happy and contented, in spite of a jealous stepmother. She was very pretty, with blue eyes and long black hair. Her skin was delicate and fair, and so she was called Snow White. Everyone was quite sure she would become very beautiful. Though her stepmother was a wicked woman, she too was very beautiful, and the magic mirror told her this every day, whenever she asked it.\\
"Mirror, mirror on the wall, who is the loveliest lady in the land?" The reply was always; "You are, your Majesty," until the dreadful day when she heard it say, "Snow White is the loveliest in the land." The stepmother was furious and, wild with jealousy, began plotting to get rid of her rival. Calling one of her trusty servants, she bribed him with a rich reward to take Snow White intc the forest, far away from the Castle. Then, unseen, he was to put her to death. The greedy servant, attracted to the reward, agreed to do this deed, and he led the innocent little girl away. However, when they came to the fatal spot, the man's courage failed him and, leaving Snow White sitting beside a tree, he mumbled an excuse and ran off. Snow White was all alone in the forest.\\
Night came, but the servant did not return. Snow White, alone in the dark forest, began to cry bitterly. She thought she could feel terrible eyes spying on her, and she heard strange sounds and rustlings that made her heart thump. At last, overcome by tiredness, she fell asleep curled under a tree.\\
Snow White slept fitfully, wakening from time to time with a start and staring into the darkness round her. Several times, she thought she felt something, or somebody touch her as she slept.\\
At last, dawn woke the forest to the song of the birds, and Snow White too, awoke. A whole world was stirring to life and the little girl was glad to see how silly her fears had been. However, the thick trees were like a wall round her, and as she tried to find out where she was, she came upon a path. She walked along it, hopefully. On she walked till she came to a clearing. There stood a strange cottage, with a tiny door, tiny windows and a tiny chimney pot. Everything about the cottage was much tinier than it ought to be. Snow White pushed the door open.\\
\\
	    \end{tabular}\\
	    \hline
	    \multicolumn{2}{|l|}{\begin{tabular}{Q}\\
	    {\bf Generated}\\
Designs concerned: Some hail the hero,\\
\\
Verdicts seem to be the wrath of the storm,\\
\\
The main reason sweets can't provide a lot of herace;\\
\\
One side calls her errible, the other side tells her,\\
\\
"Watch out, this is a great purity factory."\\
\\
"Oh, she we are degraded," a low voice whispered into her, "the delight heights thus are deeds rather than love."\\
\\
"For only things that are great, and that happiness cannot wast; For she appears so what is best.\\
\\
In this changing season, Regan might win a daily battle, In the surprise that revealing her majesty brings.\\
\\
At the end of the day, surprised and amused and bewildered," says Regan,\\
\\
"What did I wait for," demands a woman, "burst upon your beautiful mother?"\\
\\
The crease goes on and on as they argue over the choices best,\\
\\
Even though the simpler justice began to prevail, the evil abyss began to mend.\\
\\
"Such is her heart she didn't even know that,\\
\\
The mercy of her master she feared would take her life," she said, "a woman who looked only for herself.\\
\\
She pretended to think the choice that befell him was her love,But true, she turned to the figure in her bed."\\
\\
"If you like the sweetness of this sweet song," Violet sighed, "you will love them either."\\
\\
"This sweet song," says Regan with astonishing strength, "a marriage, a husband who loves her.\\
\\
This sweet song, on the one side of which is the inner figure in the solution to her woes; the sensual dictates; by which she might follow the recipe of the book of grateful dreams,\\
\\
The sweetest song I ever had ever heard, to this new flourishes. Perfect!" she said to Violet, "not just found but realised it had mean\\
\\
	    \end{tabular}}\\
	    \hline
	    
	\end{tabular}
	\end{adjustbox}
	\caption{Qualitative Results: to children books style.}
	\label{tab:app4}
\end{table}
\begin{table}[h]
\begin{adjustbox}{width=1\columnwidth,center}
	\centering
	\begin{tabular}{| l | l |}
	    \cline{1-2}
	    \begin{tabular}{U}
	    {\bf Context -- Movie Reviews Style}\\
	    I am struggling to find a good reason to give you in order to see this movie, but I cannot. I will not stick to how the actors were playing because it was not their fault about the role they had been given. On the other hand I will base my review on the scenario. What was the meaning of the movie? Which was the target of the plot? I did not manage to answer neither of the aforementioned queries. I read that\\ \\ \\ \\ \\ \\ \\ \\ \\ \\ \\ \\ \\ \\
	    \end{tabular}
	    &
	    \begin{tabular}{U}\\
	    {\bf Reference -- Politic News Style}\\
	    (CNN)President Donald Trump is wasting no time in attempting to torch House Speaker Nancy Pelosi's impeachment gamble in a battle that will define his presidency and the 2020 election.\\
	    \\
Trump has pledged to publish an un-redacted and declassified transcript on Wednesday of a phone call with Ukraine's leader at the center of what Democrats allege is his abuse of presidential power. The White House is also planning to release to Congress a whistleblower's complaint that triggered the week-long crisis that has rocked the Trump presidency.\\
Trump's decision marks a departure for a White House that has a record of obstructing oversight and bending fact. So his critics will await events on Wednesday with particular interest.\\
The transcript and the congressional reaction to the whistleblower's report could be critical in establishing the early terrain of the impeachment fight and to shaping public opinion that will ultimately dictate how it turns out.\\
Regardless of the outcome, Trump finds himself at the center of a rare and historic showdown as only the fourth president in US history to face the realistic threat of impeachment.\\
\\
	    \end{tabular}\\
	    \hline
	    \multicolumn{2}{|l|}{\begin{tabular}{Q}\\
	    {\bf Generated}\\
a short about a World War II president contemplating a secret meeting with Hitler won't be released, but the person being asked that it is, in fact, a statement he made in Munich on May 6, 1938 that would lead to the end of World War II.\\
I also assume that modern life in the present moment right now is much more uncomfortable, and it's difficult for my biographer to provide empathy because what narratives I have are so blended with existing views that the idea of Trump president is only in the realm of his childish vote to reject the Supreme Court that motivated him to do what, instead of taking a stance that is less adversarial, may be demagogic.\\
Trump by comparison isn't such a setting to gain the desired political stature that Monty Python's sketch computer shows him. Nor, it's just a new territory he's taking to be killed. Contentious and creative questions are looming. The real question is whether all this will ultimately lead to the abolition of the Supreme Court.\\
\\
	    \end{tabular}}\\
	    \hline\hline

	    \begin{tabular}{U}
	    {\bf Context -- Paper Style (Abstract of this paper)}\\
We introduce a language generative model framework for generating a styled paragraph based on a context sentence and a style reference example. The framework consists of a style encoder and a texts decoder. The style encoder extracts a style code from the reference example, and the text decoder generates texts based on the style code and the context. We propose a novel objective function to train\\ \\ \\ \\ \\ \\ \\ \\
	    \end{tabular}
	    &
	    \begin{tabular}{U}\\
	    {\bf Reference -- News Style }\\
	    (CNN)Massachusetts Gov. Charlie Baker has called for a temporary statewide ban on the sale of all e-cigarettes and vaping products in response to a nationwide outbreak of lung injuries associated with vaping.\\
	    \\
"I'm officially declaring a public health emergency in the Commonwealth due to severe lung disease associated with the use of e-cigarettes and marijuana-infused vaping products," Baker said during a press conference on Tuesday.\\
"I'm requesting that the public health council order a four-month temporary ban on the sale of all vaping products in retail establishments, online and through any other means, effective immediately," he said. "We as a Commonwealth need to pause sales in order for our medical experts to collect more information about what is driving these life-threatening vaping-related illnesses."\\
\\
	    \end{tabular}\\
	    \hline
	    \multicolumn{2}{|l|}{\begin{tabular}{Q}\\
	    {\bf Generated}\\
everyone through a useful tool or method; each custom bound to require instruction from an expert inside the business community. This type of exercise should be done in a timely manner and without fanfare and cooing from the sports, entertainment and educational communities, and in a highly professional scientist setting, where everyone could help us understand things we need to improve…we want people to be able to start thele by entering into an active conversation with scientists -- which is a "great idea," if, by definition, is humans for."\\
The medical community has an primary role in the global health and well-being of the globe beyond infections, obesity, heart disease, diabetes, and opioid addiction," Bolton added. "This includes controlling just-induced drugs. But as a potent encapsulation of these drugs, we cannot defend their efficacy for all measures, including food. Simply creating something that doesn't currently prohibit prescription drugs, and packaging them for adult use, is wholly irresponsible on both sides of the political aisle, our ability to regulate the reflux to the world's thirsty, and the health of the Australian people."\\
\\
	    \end{tabular}}\\
	    \hline\hline
	    \begin{tabular}{U}
	    {\bf Context -- Children Books Style (Little Three Pigs)}\\
Once upon a time there was an old mother pig who had three little pigs and not enough food to feed them. So when they were old enough, she sent them out into the world to seek their fortunes.\\
\\
The first little pig was very lazy. He didn't want to work at all and he built his house out of straw. The second little pig worked a little bit harder but he was somewhat lazy too and he built his house out of sticks. Then, they sang and danced and played together the rest of the day.\\
\\
The third little pig worked hard all day and built\\ \\ \\ \\ \\ \\ \\ \\ \\ \\ \\ \\
	    \end{tabular}
	    &
	    \begin{tabular}{U}\\
	    {\bf Reference -- News Style }\\
	    (CNN)For Quentin Tarantino, time travel is entirely possible. And it didn't take green screens or computer generated backdrops to make it back to 1969-era Los Angeles. All it required were fond, vivid memories of the time, an exhausting attention to period detail and a production designer like Barbara Ling.\\
	    \\
The filmmaker was determined that the environments featured in his ninth film "Once Upon a Time ... in Hollywood" as tactile and immersive as humanly possible without turning to the now-ubiquitous digital effects houses, says Ling, who lavishly and convincingly resurrects a period some 50 years gone: the then-vibrant storefronts of Hollywood Boulevard, the classic movie theaters of Westwood Village, enduring eateries like Musso \& Frank Grill and El Coyote, the Playboy Mansion, and the Charles Manson family inhabited ranch where Westerns were shot.\\
\\
"Quentin is so not into digital or CGI," Ling, whose film credits include "The Doors," "Falling Down" and "Batman Forever," told CNN. "He's not someone who wants to do performance in front of green screen. He wants to see the real deal. To me, that is, of course, the absolute best way. It's the hardest way, but you actually get a real 3-D world that's not hazy, so to speak, or that you have to keep people at a certain distance. You're not dealing with green screens. You're dealing with real streets and real people and real everything...He loves the real world, so you can actually see it and smell it and touch it."\\
\\
	    \end{tabular}\\
	    \hline
	    \multicolumn{2}{|l|}{\begin{tabular}{Q}\\
	    {\bf Generated}\\
a new house,\\
away from that stage, away from the theater,\\
away from the movie.\\
\\
But the fourth little pig worked hard — it had to be him.\\
\\
Steven Spielberg is one of us white people,
whom we think of as "the lovable Dear Dad".\\
\\
In the end, Spielberg created the "Ralph" sequence with Donkey Kong.\\
\\
Everyone who adopts a culture of old blood, that's Abraham Lincoln for those of you white people.\\
\\
I once invited a school animal to a movie so that I could ask him, 'Maybe Jaws + Shaun of the Dead', forever. She said, 'Sure.' 'I'll be your film, but I don't want to be one of you gum and crackers any more.' She's this confused bunny."\\
\\
(Click here to see a video line-up of Teri Garr's past "Ralph" scenes and upcoming "Dreaming Tom Shadow.")\\
\\
	    \end{tabular}}\\
	    \hline
	    
	\end{tabular}
	\end{adjustbox}
	\caption{Qualitative Results: to News style.}
	\label{tab:app5}
\end{table}

\end{document}